\documentclass[letterpaper, 10 pt, journal, twoside]{IEEEtran}
\ifCLASSINFOpdf
\else
\fi

\newcommand{\figscale}{1}

\usepackage{amsmath, amssymb, amsthm}  
\usepackage{color, xcolor}            
\usepackage{float}                    
\usepackage{graphicx}                 

\usepackage{algorithm}
\usepackage{algorithmic}
\usepackage{booktabs}                 
\usepackage{caption}                  
\usepackage{diagbox}                  
\usepackage{multirow}                 
\usepackage{subfigure}                

\usepackage{enumitem}                 
\usepackage{lipsum}                   
\usepackage{soul}                     

\usepackage{acro}                     
\usepackage{cite}                     
\usepackage{hyperref}                 

\DeclareAcronym{DLAS}{
	short = DLAS ,
	long = Different Length Alignment Sewing
}

\DeclareAcronym{DLFFSC}{
	short = DLFSC ,
	long = Different Length Feeding Speed Control
}

\DeclareAcronym{DLRoSS}{
	short = DLRoSS ,
	long = Different Length Robotic Sewing System
}

\DeclareAcronym{RoSS}{
	short = RoSS ,
	long = Robotic Sewing System
}

\DeclareAcronym{FFRoSS}{
	short = FFRoSS ,
	long = Fixed Fabric Robotic Sewing System
}

\DeclareAcronym{NFRoSS}{
	short = NFRoSS ,
	long = Non-fixed Fabric Robotic Sewing System
}

\DeclareAcronym{FST}{
	short = FST ,
	long = Fabric Stabilizing Tool
}

\begin{document}
\title{Automated Straight-line Sewing of Stretchable Fabrics with Different Lengths}
\author{%
 	Bingchen Jin\textsuperscript{$\dagger$1},
 	Akinari Kobayashi\textsuperscript{$\dagger$1},
 	Dipankar Bhattacharya$^{1}$,
	Akira Seino$^{1}$,
 	Fuyuki Tokuda$^{1}$,\protect\\
 	Norman Chihnan Tien$^{2}$,
 	and Kazuhiro Kosuge$^{1}$%
    
\thanks{Manuscript received: August, 19, 2025; Accepted September, 26, 2025.}
\thanks{This paper was recommended for publication by Editor Gosselin, Clement upon evaluation of the Associate Editor and Reviewers' comments.
This work was supported in part by the Innovation and Technology Commission of the HKSAR Government under the InnoHK initiative. The research was conducted in part at the JC STEM Lab of Robotics for Soft Materials, funded by The Hong Kong Jockey Club Charities Trust.} 
\thanks{$^{1}$B. Jin, A. Kobayashi, D. Bhattacharya, A. Seino, F. Tokuda,  and K. Kosuge are with the JC STEM Lab of Robotics for Soft Materials, Department of Electrical and Electronic Engineering, Faculty of Engineering, The University of Hong Kong, Hong Kong SAR, China, and are with the Centre for Transformative Garment Production, Hong Kong Science Park, Hong Kong SAR, China. (e-mail: bcjin@eee.hku.hk; akinari.kobayashi@transgp.hk; d.bhattacharya@transgp.hk; akira.seino@transgp.hk; fuyuki.tokuda@transgp.hk; kosuge@hku.hk).}
\thanks{$^{2} $N. C. Tien is with the Department of Electrical and Electronic Engineering, Faculty of Engineering, The University of Hong Kong, Hong Kong SAR, China, and is with the Centre for Transformative Garment Production, Hong Kong Science Park, Hong Kong SAR, China (e-mail: nctien@hku.hk).}%
\thanks{\textsuperscript{$\dagger$}B. Jin and A. Kobayashi are co-first authors.}
\thanks{Digital Object Identifier (DOI): see top of this page.}
\thanks{\copyright~2025 IEEE. Personal use of this material is permitted. Permission from IEEE must be obtained for all other uses, in any current or future media, including reprinting/republishing this material for advertising or promotional purposes, creating new collective works, for resale or redistribution to servers or lists, or reuse of any copyrighted component of this work in other works.}
}

\markboth{IEEE Robotics and Automation Letters. Preprint Version. Accepted September, 2025}
{Jin \MakeLowercase{\textit{et al.}}: Automated Straight-line Sewing of Stretchable Fabrics with Different Lengths} 

\maketitle
\begin{abstract}
\ac{DLAS}, which involves stretching the shorter fabric to match the longer one and sewing them together in a straight line, is a challenging task that needs to satisfy several requirements when automating the sewing process.
To address the challenges, this research proposes a novel robotic sewing system, \ac{DLRoSS}, which consists of a roller type end-effector, attached to a 6-DoF manipulator. The end-effector composed of active shorter and longer fabric rollers, and a passive press-roller attached to the shorter-fabric roller. Assuming that one end of the two fabric layers are initially positioned under the sewing machine's presser foot, the system automates DLAS by operating in four distinct phases. (P1) Fabric wrapping: Individual fabric layers are picked, held, and wrapped from the other end onto the feed rollers. (P2) Sewing: During the sewing, the shorter fabric is stretched and aligned with the longer fabric in real-time using roller velocity control based on the sewing speed and apriori known length ratio. (P3) Sewing completion: In the final sewing round on the fabric rollers, the press roller is engaged to prevent the stretched fabric from slipping off due to internal tension. (P4) Sewing fabric release: At the end of sewing, the fabric edge moves past the press roller, and the fabric releases from the rollers.
Experimental results demonstrate that \ac{DLRoSS} achieves consistent, high-quality sewing of stretchable fabrics of different materials and lengths.

\end{abstract}


\IEEEpeerreviewmaketitle

\acresetall
\section{Introduction}
\label{sec_introduction}

\IEEEPARstart{S}{ewing} is the most crucial process in the garment production, as it involves approximately 80\% of the fabric stitching and accounts for 40\% of the total production cost, making it one of the most resource-intensive and labor-intensive processes \cite{gries2018application}. 
The sewing process becomes more intricate and time-intensive in  \textit{\ac{DLAS}}, where two fabrics of different lengths must be aligned and sewn together along a straight line by stretching the shorter fabric to match the longer one.
This operation is common in garment manufacturing such as attaching elastic bands to shirts and jackets. Hence, automated sewing systems, particularly for \ac{DLAS}, are essential for improving time and cost efficiency, ultimately boosting productivity. 

\begin{figure}[t!]
	\centering
	\includegraphics[width=\figscale\linewidth]{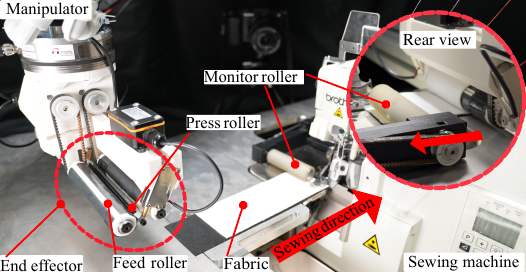} 
	\caption{Proposed robotic sewing system for DLAS.}
	\label{fig:dfross} 
	\vspace{-1em} 
\end{figure}

Designing an automated system for DLAS requires handling multiple fabric layers with independent tension control while allowing deformation during sewing. It must also grasp and separate stacked fabrics automatically, as assuming pre-grasped materials limits applicability. 
The end-effector should hold fabrics securely without displacement, adapt to stretchable textiles, and prevent issues such as detachment from rollers caused by stretching tension. Precise coordination between rollers, manipulators, and the sewing machine is essential to maintain synchronization and achieve accuracy within garment industry tolerances \cite{JIS,9551585}. Overall, the system must integrate grasping, stretching, handling, and sewing into a unified workflow that overcomes the shortcomings of fixture-based, fixture-free, and roller-based approaches.

Traditional automated sewing systems can be grouped into two main types: \textit{fixture-based systems} and \textit{fixture-free systems}. Fixture-based systems, commonly used in semi-automated solutions such as pattern sewers and pocket setters, use mechanical fixtures or templates to secure and guide fabrics during sewing~\cite{ku2023automated}. However, as these fixtures restrict fabric deformation, this approach is unsuitable for DLAS. Current fixture-free systems also face limitations for DLAS automation. Most of the systems are designed for single-layer fabric and cannot independently control multiple layers, which is essential for DLAS \cite{gershon1990parallel}. While some use multiple manipulators and grippers, they typically assume fabrics are pre-grasped, without addressing the complex task of separating and grasping stacked fabrics \cite{schrimpf2013real}. Roller-based end-effectors, though adaptable, often struggle to consistently hold stretched fabrics, affecting sewing accuracy \cite{9551585}. Furthermore, many methods treat fabric as a rigid body \cite{tang2025fixturefree}, failing to accommodate the stretchability required in DLAS. Consequently, none of the current automated sewing systems are directly applicable to DLAS whose automation remains relatively unexplored.

Hence, this research proposes a novel automated sewing system, \textit{\ac{DLRoSS}}, which sews two stretchable fabrics of different lengths and materials along a straight line without relying on fixtures or templates, as shown in Fig.~\ref{fig:dfross}.
The main contributions of this paper are as follows
\begin{itemize}
	\item \textcolor{black}{To the best of our knowledge, this is the first robotic sewing system proposed to control fabric deformation to automatically stretch and sew together two fabrics of different lengths.}
	\item The proposed system consists of: 1) a novel roller-based end-effector, 2) a novel actual fabric feeding speed sensing system, 3) a manipulator, and 4) an overlock sewing machine, driven by a servo motor.
    The manipulator and end-effector roll the fabric onto a roller before sewing to maintain the desired feeding direction and prevent gradual deviation.
    To prevent fabric slippage during stretching, the end-effector uses a press-roller, akin to sewing machine presser feet,  which is a novel design. 
    The sensing system measures the actual fabric feeding speed based on a roller-based mechanism for precise DLAS.

    \item \ac{DLRoSS} experimental results show that the proposed system performs DLAS precisely and successfully for fabrics of different materials and lengths.
\end{itemize}

The remainder of the paper is organized as follows. The related work is reviewed in Section II. Section III formulates the DLAS operation phases and advantages. Section IV details the DLRoSS construction, working principles, and control framework. Section V provides the experimental results, and Section VI concludes the paper with future directions.

\section{Related work}
\label{sec_related_work}

\subsection{Fixture-based systems}

Fixture-based systems are widely used in semi-automated sewing systems in the market such as pattern sewers
and pocket setter, and use mechanical fixtures or templates to secure and guide fabrics during sewing \cite{ku2023automated}. Ogura et al. \cite{ogura2022automation} used mechanical guides to sew rectangular and semi-circlar fabrics. 
However, these fixtures, templates, or mechanical
guides need customization for the shape, size, and material of each pair of fabrics. Also, restricted by the fixture, the fabric is usually unable to be deformed, which makes this method unsuitable for DLAS.

\subsection{Fixture-free systems }
Fixture-free systems are designed to perform sewing fabrics of different shape, size, and material without using mechanical fixtures or templates.
Gershon et al. \cite{gershon1990parallel,gershon1988Vision} realized automated sewing by dividing it into four parallel processes: observing the fabric edge, modifying tension, controlling robot feed motion, and controlling the sewing machine. 
Winck et al. \cite{winck2009novel} proposed a novel fabric manipulation method for controlling the fabric sewing direction using a novel sewing machine feeddog mechanism.
Tokuda et al. \cite{tokuda2023fixture} proposed a 2D fixture-free sewing system using a dual-arm manipulator system to sew stacked fabrics along the desired seam line. 
Tang et al. \cite{tang2024time,tang2025fixturefree} also proposed a 2D sewing method by modeling the fabric movement as a non-holonomic system.
Further, Zacharia et al. \cite{Zacharia2010Neurofuzzy,Zacharia2008robot} proposed a method to detect the fabric's edge and thus control the fabric's sewing direction based on visual servoing and a learning technique that combines neural networks and fuzzy logic.
\textcolor{black}{While these platforms can effectively control the sewing direction, the designed end-effector or specialised sewing machine modification limits their ability to control and stretch each layer of fabric individually, making these platforms difficult for performing DLAS.} 

Some research uses grippers to control layers of fabric individually.
Schrimpf et al. proposed a sewing cell in his series of studies that using multiple robot arms equipped with specially designed grippers to handle the fabrics by controlling their sewing direction\cite{schrimpf2013real,schrimpf2015model,schrimpf2012real,Schrimpf2012Experiments,Schrimpf2014Velocity}.
Kudo et al. \cite{kudo2000multi} proposed an automated sewing system consisting of two manipulators to handle the fabric on the sewing table. With multi-sensors, this system can perform fabric tension compensation, sewing machine speed synchronization and collision avoidance during sewing task execution.
Triantafyllou et al. introduced a control method to regulate the fabric's tension force during sewing using fuzzy model \cite{triantafyllou2011model,panagiotis2014intelligent}.
\textcolor{black}{Although these systems achieved automatic sewing and addressed the challenge of independent layer manipulation, they did not propose a method for automatically grasping stacked fabrics, instead assuming the fabrics were already pre-grasped, a step that remains a significant challenge in automated sewing due to the lack of an automated fabric grasping method.}

\textcolor{black}{To overcome the challenge of automatically grasping fabrics from a stack, more recent approaches have used roller-based end-effectors, which have demonstrated high adaptability for automatic sewing.
}
Tajima et al. \cite{9551585} proposed a dual-arm manipulator with roller end-effectors to hold down fabric during manipulation for automatic feeding during sewing.
Kosaka et al. proposed an automated sewing system using specially designed robotic arms and implemented a control system for 3D sewing \cite{kosaka2023real}. 
\textcolor{black}{Although the roller makes it easier to control both the fabric sewing direction and its feeding speed, the platforms proposed in their research are still only suitable for sewing one layer of fabric. Also, for roller-based end-effector, it was frequently observed that the stretched fabric would fall off the roller during our primary DLAS experiment whose reason will be explained in Sec. IV. Although this issue has not been discussed in previous automated sewing research, failing to address it could significantly reduce sewing accuracy.}

\textcolor{black}{Apart from that, related research \cite{schrimpf2013real,tang2025fixturefree} tends to ignore the stretchability of fabric, treating it as a rigid body, and utilizing force sensors to guide the fabric feeding speed which makes their method only applicable to non-stretchable fabrics, thus failing to complete DLAS.}
\textcolor{black}{In summary, none of the related automatic sewing systems can be applied for \ac{DLAS} automation directly.}

\section{System overview and operation phases}
\label{sec_problem-formulation}

This section formulates the \ac{DLAS} problem and introduces the proposed \ac{DLRoSS}, detailing its key components, and operation phases. It also highlights the functional advantages of DLRoSS in addressing the menthioned limitations.

\begin{figure}[t!]
	\centering
	\includegraphics[width=\linewidth]{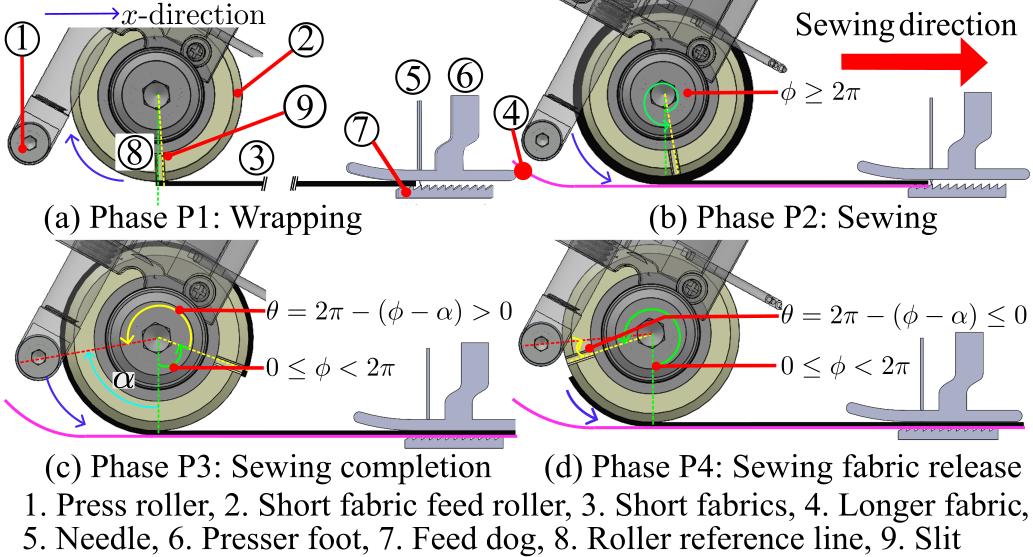} 
	\caption{ DLRoSS operation phases and fabric conditions. 
}
	\label{fig:dfas_phases} 
	\vspace{-1em}
\end{figure}

\subsection{Proposed system overview}
Given two fabrics of different lengths, \( l_s \) (shorter) and \( l_l \) (longer), aligned and fixed at one end under the sewing machine’s presser foot, this work proposes a \ac{DLRoSS} (Fig.~\ref{fig:dfross}) for performing straight-line \ac{DLAS}.
The major components of the DLRoSS are as follows: 
\begin{enumerate}
    \item An \textit{end-effector} for fabric manipulation comprises \textit{two active feed rollers} for manipulating the shorter and longer fabrics, referred to as the \textit{shorter fabric feed roller} and the \textit{longer fabric feed roller}, respectively. Additionally, a \textit{passive press roller} is attached to the shorter fabric roller to prevent fabric slippage from the end-effector (Sec.~\ref{sec:end_effector}).
    \item A \textit{fabric feeding speed sensing system} comprising a \textit{monitor roller} for measuring the actual fabric feeding speed to the sewing machine (Sec.~\ref{sec:monitor_roller}).
    \item A manipulator for moving the end-effector to the desired fabric location (Sec.~\ref{sec:other_hardware}).
    \item An overlock sewing machine for sewing the two fabrics (Sec.~\ref{sec:monitor_roller}).
\end{enumerate}
\subsection{DLRoSS operation phases and advantages}

To explain the DLRoSS operation, several auxiliary lines and angles are defined.
The \textit{roller reference line} is defined as the line perpendicular to the plane at the center of the shorter fabric feed roller, illustrated by the green dashed line in Fig.~\ref{fig:dfas_phases}. The angle $\phi$ is defined as the angle between the roller reference line and the line joining the center of the roller to its slit (yellow dash line). Similarly, $\alpha$ is the angle between the roller reference line and the line joining the roller center to the press roller center when it is activated (red dash line). Based on the $\phi$ and $\alpha$, the DLRoSS operation can be divided into four primary phases:
\begin{enumerate}
   \item \textbf{Phase P1: Fabric wrapping} (Fig.~\ref{fig:dfas_phases} (a)) -- The \ac{DLRoSS} manipulator aligns the feed rollers' slit with the fabric end-effector side. Using an \textit{air suction mechanism} through a slit (Sec. \ref{sec:end_effector}), the rollers pick and hold each fabric layer, rolling excess fabric \textit{clockwise} while the manipulator moves the rollers with a linear velocity \( v_x \) in $x$-direction.
	
	\item \textbf{Phase P2: Sewing}, when $\phi\ge2\pi$ (Fig.~\ref{fig:dfas_phases} (b)) -- The sewing starts, while the manipulator is stationary ($v_x=0$), the presser foot secures the fabric and the feed dog moves it forward during sewing, as the fabric is unwrapped and aligned by stretching from the rollers in an \textit{anticlockwise} movement. The feed roller velocity for the shorter fabric, \( v_s(t) \), is controlled using the estimated sewing speed, \( v_{sew}(t) \), measured by a \textit{monitor roller}-based sensing system (Sec. \ref{sec:monitor_roller}). This control stretches the fabric according to the \textit{sewing length ratio} \( \rho = {l_s^{sew}}/{l_l^{sew}} \), where \( l_s^{sew} \) and \( l_l^{sew} \) denote the sewing lengths of the shorter and longer fabrics, respectively. Simultaneously, the feed roller velocity for the longer fabric, \( v_l(t) \), is matched with \( v_{sew}(t) \) to prevent slack.

	\item \textbf{Phase P3: Sewing completion}, when $0\le \phi < 2\pi$ and $ \phi - \alpha < 2\pi$ (Fig.~\ref{fig:dfas_phases} (c)) -- This phase involves the final round of unwrapping and sewing. Based on the feed roller configuration in this research, the stretched fabric will always fall off the roller in this phase, the reason for which will be explained in the next section. To solve this, the press roller (Sec. \ref{sec:end_effector}), positioned next to the feed roller, is pressed firmly (retracted) against the shorter fabric feed roller to prevent fabric slippage caused by the fabric's internal tension.
	
	\item \textbf{Phase P4: Sewing fabric release}, when $0\le \phi < 2\pi$ and $ \phi-\alpha \ge 2\pi$ (Fig.~\ref{fig:dfas_phases} (d)) -- 
    Toward the end of the sewing process, the shorter fabric edge passes the pressing point of the press roller. At this point, the air suction hold becomes insufficient to resist the fabric’s internal tension, causing the fabric to disengage and fall off the roller, which results in loss of fabric hold by the end-effector.
\end{enumerate}

The DLRoSS design offers four main advantages.
\textcolor{black}{First, it enables the end-effector to wrap and pick the entire fabric before sewing, eliminating the need to pause and re-grasp the fabric during the sewing process.
Second, utilizing two feed rollers, the two fabric pieces can be controlled individually.
Third, placed beside the sewing machine's presser foot, the monitor roller can observe the actual sewing speed without being influenced by the fabric's deformation.
Finally, the press roller mechanism solve the fabric slippage which will always occur for roller-type end-effector in \ac{DLAS}.}

\section{DLRoSS design and control}
\begin{figure*}[!h]
	\centering
	\includegraphics[width=\textwidth]{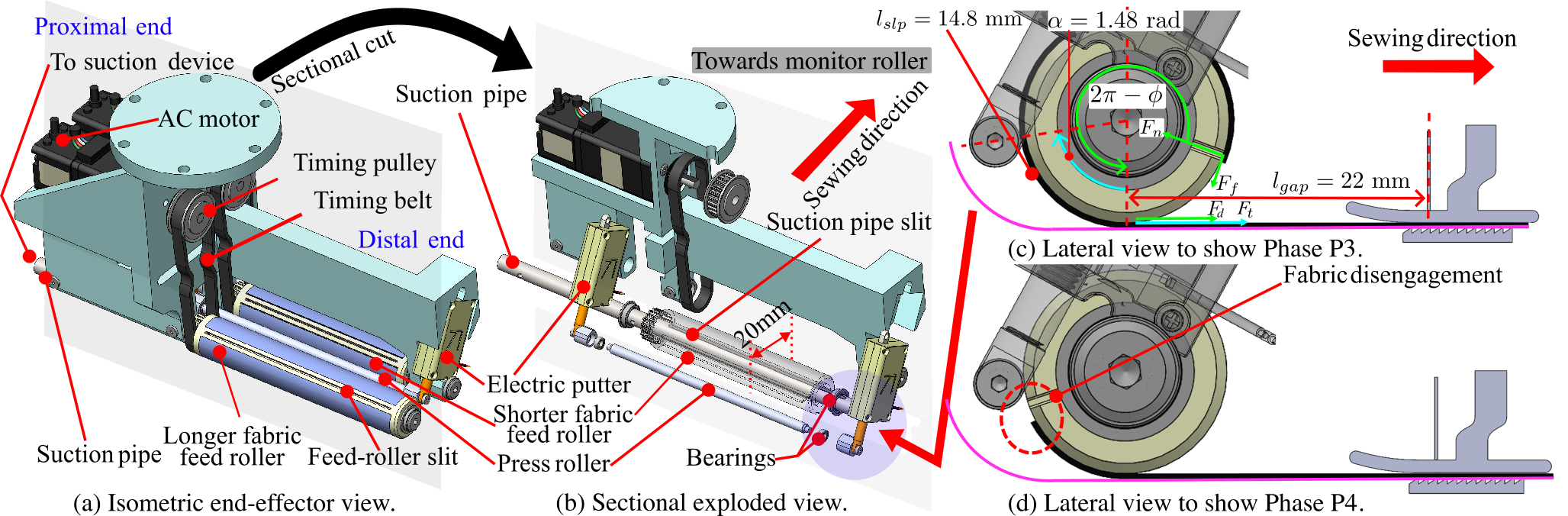} 
	\caption{Construction of the DLRoSS end-effector and its operation during phases P3–P4.} 
	\label{fig:end-effector} 
	\vspace{-1em} 
\end{figure*}
This section provides a comprehensive discussion on the construction and working principles of each \ac{DLRoSS} components and its control system for sewing speed estimation, and feed roller coordination. 
\subsection{The End-effector}\label{sec:end_effector}
The designed end-effector (Fig.~\ref{fig:end-effector}) can be divided into two major parts according to its function: the active feed rollers, and the passive press roller. 
\paragraph*{Active feed-rollers}
A pair of hollow steel suction pipes, with axial slits 2 mm wide and 108 mm long for air suction, are inserted through the end-effector base and extend to the other side, with the slits aligned coincident with the sewing table (Fig.~\ref{fig:end-effector}) (a)--(b)). The proximal ends of the pipes are connected to separate suction devices, while their distal ends fully house two feed rollers, a longer and a shorter fabric roller, with diameters of 20 mm and lengths of 120 mm and 100 mm, respectively, each featuring similar axial slits, measuring 2 mm in width and 108 mm and 88 mm in length, respectively for air suction.

These rollers are connected to the pipes through ball bearings placed circumferentially inside the rollers at both the proximal and distal ends, allowing them to rotate freely. This design enables the rollers to pick and wrap two layers of fabric from stacked fabrics during Phase P1 and feed the fabric into the sewing machine during the sewing process.

Further, the outer side of the proximal end of the feed rollers is shaped as a timing pulley so that they can be driven by AC servo motors (Yaskawa SGM7M) through timing belts and timing pulleys. To better interact with the fabrics, silicone skins (Formlabs Silicone 40A resin) are attached to the surface of rollers to increase the friction coefficient.

\paragraph*{Passive press-roller}

During Phase P3 (Fig.~\ref{fig:dfas_phases} (c), and Fig.~\ref{fig:end-effector}), let \( F_{t} \) denote the fabric tension force and \( \mu \) the static friction coefficient between the roller and the fabric. Then, the \textcolor{black}{\textit{desired fabric holding force} \( F_{d} \), which needs to be applied on the short fabric to hold its edge in the feed roller slot, is given by the following capstan equation}
\begin{equation}
	\begin{aligned}
		&F_{d} = F_{t}e^{-\mu\left(2\pi - \phi\right)}.
	\end{aligned}
	\label{eq:1}
\end{equation}
\textcolor{black}{Next, the desired fabric holding force $F_d$ is met by the friction force between the fabric and the feed roller, whose maximum value can be expressed as}
\begin{equation}
	\begin{aligned}
		&F_{f} = \mu F_n, 
	\end{aligned}
	\label{eq:2}
\end{equation}
\textcolor{black}{where \( F_n \) represents the normal force exerted by the suction mechanism between the roller and the fabric at the slit.}

To keep the fabric on the slit and avoid fabric slippage, it is necessary that \( F_{f} \geq F_{d} \). However, as the \ac{DLAS} progresses, \( \phi \) increases, causing \( F_{d} \) to rise. Eventually, \( F_{d} \) exceeds \( F_{f} \), resulting in the fabric slipping off the roller and disengaging from the slit. Based on the configuration of the feed rollers, specifically the nominal torque of the motor and the friction coefficient of the roller's surface, this phenomenon occurs when the fabric wraps around the roller in fewer than one complete circle (P3).

To address this, an aluminum press roller (Fig.~\ref{fig:end-effector} (a)--(b)) with a diameter of 5 mm and a length of 118 mm is attached adjacent to the shorter fabric feed roller. Similar to the feed rollers, the press roller is supported by bearings at both ends and is linearly actuated by an electric putter at each end, with a stroke of 5 mm. This roller is controlled by an I/O signal \( s_p \), which commands its extension and retraction via electric putters on both sides. During Phases P1 and P2, the electric putter controlling the press roller remains fully extended, maintaining a maximum gap of 3.3 mm between the press roller and the feed roller, and 5.0 mm between the press roller and the sewing table, which ensures that the press roller does not interfere with the sewing environment. 

During P3, the press roller (Fig.~\ref{fig:end-effector} (c)) applies pressure on the fabric, increasing the normal force \( F_n \) in (\ref{eq:2}). This raises the friction force \( F_f \) between the feed roller and fabric, ensuring \( F_f \geq F_d \) so the fabric stays fixed on the slit until P4, when it disengages and falls off (Fig.~\ref{fig:end-effector}(d)). \textcolor{black}{Although passive rollers are common in sewing machines as presser feet, their use in end-effector designs to prevent fabric slippage during stretching is an innovative application.}

\vspace{-1em}
\subsection{Fabric feeding speed sensing}
\label{sec:monitor_roller}

\begin{figure}[t!]
	\centering
	\includegraphics[width=\figscale\linewidth]{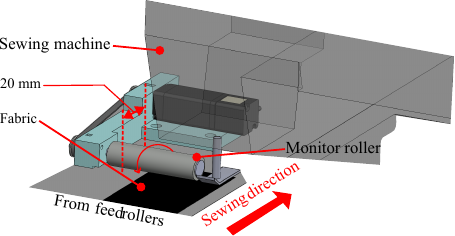} 
	\caption{Monitor roller design for sewing speed estimation. }
	\label{fig:monitor_roller} 
    \vspace{-1em} 
\end{figure}

\textcolor{black}{A straightforward method to sense the actual sewing speed involves placing a passive wheel with an encoder beneath or beside the presser foot. However, the friction generated by the sensing system would stretch the fabric, effectively turning the system into a payload rather than a speed sensor. This unintended interaction may lead to fabric distortion, including wrinkling, during the sewing process.} 
Hence, in this research, a 20\,mm torque-controlled monitor roller (Fig.~\ref{fig:monitor_roller}) is attached to enable real-time sewing speed sensing during Phases P2 and P3. Its surface is coated with polyurethane to increase friction and prevent sliding relative to the fabric. The roller is driven by an AC servo motor (Yaskawa SGM7M) via a timing belt and pulleys, and the motor’s encoder measures the monitor roller's angular velocity $v_m(t)$.

When the fabric is pressed under the presser foot, it is simultaneously pressed by the monitor roller, and during sewing, it is desired that as the fabric is fed into the sewing machine, the roller rotates synchronously with the sewn fabric movement, providing real-time sewing speed estimation. 

\vspace{-1em}
\subsection{Other Hardware}\label{sec:other_hardware}

The \ac{DLRoSS} also includes an industrial overlock sewing machine, 6-DoF manipulator, and a control workstation. These components are interconnected via EtherCAT, ensuring fast and reliable operation. To ensure reliable and stable communication, INtime is used to build the EtherCAT master, enabling the system to operate as a high-frequency, real-time network.

\paragraph*{Over clock sewing Machine} 
The industrial overlock sewing machine (Brother FB-N21A) is mounted on a sewing table together with an aluminum plate within an aluminum frame. To simplify the communication system, the original motor is replaced with an AC motor (Yaskawa SGM7A), which is controlled by a servo (Yaskawa SGD7S) connected to the control workstation. 

\paragraph*{Manipulator}  
To move the roller end-effector, a 6-DoF manipulator (Denso VS-068) is suspended from an aluminum frame, with the designed end-effector attached to it via a wrist force/torque (F/T) sensor (ATI Axia80-M8). \textcolor{black}{Although sewing involves minimal movement, the manipulator remains essential. By rolling the fabric onto a roller before sewing, the manipulator and end-effector maintain the correct feeding direction, preventing gradual misalignment and ensuring consistent sewing quality.}

\vspace{-1em}
\subsection{\acf{DLFFSC}}
\label{sec:dlffsc}
\begin{figure}[t!]
	\centering
	\includegraphics[width=\columnwidth]{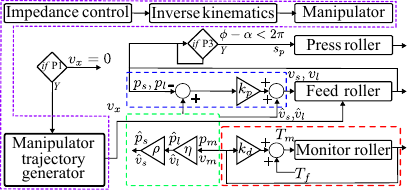}
	\caption{\acf{DLFFSC}.}
	\vspace{-1em}
	\label{fig:control}
\end{figure}

To operate the \ac{DLRoSS}, a corresponding control scheme is proposed, as shown in Fig.~\ref{fig:control}, and is explained in this section with respect to the different \ac{DLRoSS} operation phases. \textcolor{black}{The operation phases can complete DLAS by stretching and sewing the fabrics and address the fabric slipping phenomena that have never been considered.}

\paragraph*{Phase P1 control} Before sewing, the manipulator moves the end-effector, whose control scheme is shown by the purple-dashed box in Fig.~\ref{fig:control}. During Phase P1, its position is planned by defining an endpoint trajectory with linear velocity toward the sewing machine. To ensure safe interaction with the platform, impedance control \cite{648976,10196184} is applied to avoid excessive pressure that could damage components. Based on the planned trajectory and control strategy, the manipulator's trajectory is converted into joint velocities via inverse kinematics to control motion.

\paragraph*{Phase P2--P3 control}

During the sewing phase, while the manipulator remains stationary ($v_x=0$), the velocities of the \ac{DLRoSS} end-effector's shorter and longer feed rollers, \(v_s\) and \(v_l\), must be determined accurately to ensure fabric stretching and alignment in \ac{DLAS}.  Ideally, the velocities can be estimated (green-dashed box, Fig. \ref{fig:control}) from the monitor roller's angular velocity $v_m$, and are denoted as  \(\hat{v}_s\) and \(\hat{v}_l\), respectively, which can be written as

\begin{equation}
	\label{eq:feed_roller_vel_ideal}
	\hat{v}_{s}(t)\ = 	\rho \eta {v}_{m}(t) , \quad 	\hat{v}_{l}(t) = \eta v_{m}(t) ,
\end{equation}
where
\begin{equation}
	\label{eq:rho}
	\eta = \frac{r_m}{r_f}, \quad \rho=\frac{l_s^{sew}}{l_l^{sew}},
\end{equation}
where \(\eta\) is the ratio of the radii of the monitor roller to the feed roller, with \(r_m\) and \(r_f\) denoting the radii of the monitor roller and the feed roller, respectively, and  \(\rho \in (0, 1]\) is sewing fabric length ratio.

In practice, the estimation given by \eqref{eq:feed_roller_vel_ideal} significantly deteriorates due to the noise present in $v_m$, which consequently, reduces the performance of the \ac{DLAS}. 
To improve the estimation, a proportional controller (blue-dashed box, Fig. \ref{fig:control}) for each feed roller is implemented using the monitor roller's position, \(p_m(t)\). The control law is given as 
\begin{align}
	\label{eq:feed_roller_vel_comp}
	\begin{bmatrix}
		v_{s}(t)\\
		v_{l}(t)
	\end{bmatrix}
	&=
	\begin{bmatrix}
		\hat{v}_{s}(t) + k_p \left( \rho \eta p_{m}(t) - p_{s}(t) \right) \\
		\hat{v}_{l}(t)  + k_p \left( \eta p_{m}(t) - p_{l}(t) \right)
	\end{bmatrix},
\end{align}
where $k_p$ is the proportional gain and $p_s(t)$ and $p_l(t)$ are the rotation position of two feed rollers individually.

\paragraph*{Phase P4 loss of control}As the fabric transitions from Phase P3 to P4, the \textit{uncontrollable fabric length} \( l_u = l_{slp} + l_{gap} \) is the sum of the distance between the sewing machine's needle and the vertical axis of the shorter feed roller \( l_{gap} \) and the arc length from the feed roller's vertical axis to the press roller's axis \( l_{slp} = \alpha r_f \), where \( \alpha \) and \( r_f \) denote the press roller's angular position and the feed roller's radius, respectively (Fig.~\ref{fig:end-effector} (c)).
This uncontrollability arises because the \ac{DLFFSC} loses control of the shorter fabric feed roller once the fabric disengages.
Hence, instead of aligning the fabric edges, the \ac{DLFFSC} is commanded to align the fabric at uncontrollable fabric length $l_u$ distance from the end-effector side fabric edge (Fig.~\ref{fig:uncontrollable_length} (a)--(b)). Taking $l_u$ into considerations, the fabric sewing lengths ratio can be written as
\begin{equation}
	\label{eq:rho_updated}
	\rho = \frac{l_s^{sew}}{l_l^{sew}}=\frac{l_s -l_u}{l_l -l_u}, 
\end{equation}
where \(\rho \) indicates, from distance \( l_u \), how much the shorter fabric must be stretched to match the longer one.

\subsection{Monitor Roller Torque Control}
Since the feeding of the sewing machine is intermittent action, applying a constant torque to the monitor roller can cause it to slide relative to the sewn fabric since the roller alternates between starting and stopping during sewing. 
Hence, \textit{torque control} (red-dashed box, Fig. \ref{fig:control}) is employed to regulate the monitor roller’s motion, using a relatively small torque sufficient to overcome the mechanical transmission friction. 
The control law governing the monitor roller is defined as
\begin{equation}
	T_m(t) = T_{f} + k_d v_m(t),
	\label{eq:motor_torque}
\end{equation}
where \(T_m(t)\) is the monitor roller's motor torque command, \(T_{f}\) is the \textit{experimentally measured torque} required to overcome {mechanical static friction}, and \(k_d\) is the damping coefficient. The damping term in \eqref{eq:motor_torque} automatically reduces the torque command, when the monitor roller rotates, aligning with the actual physical behavior. This adjustment eliminates slippage, ensuring smooth and wrinkle-free sewing, while performing real-time sewing speed sensing.

\section{Experiments}
This section presents \ac{DLRoSS} experimental results, focusing on sewing speed estimation accuracy, the impact of the press roller on fabric alignment, and system robustness across various fabric materials and lengths.
\begin{figure}[t!]
	\centering
	\includegraphics[width=\figscale\columnwidth]{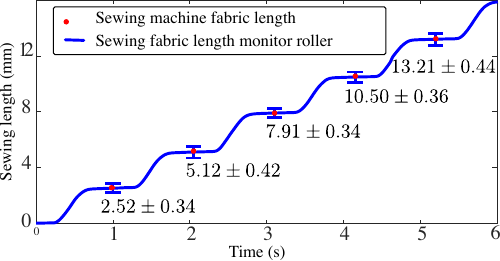}
	\caption{Monitor roller sewing speed validation.}
	\vspace{-1em}
	\label{fig:validating_monitor_roller}
\end{figure}
\subsection{Validating monitor roller sewing speed }
\begin{figure*}[t!]
	\centering
	\includegraphics[width=\textwidth]{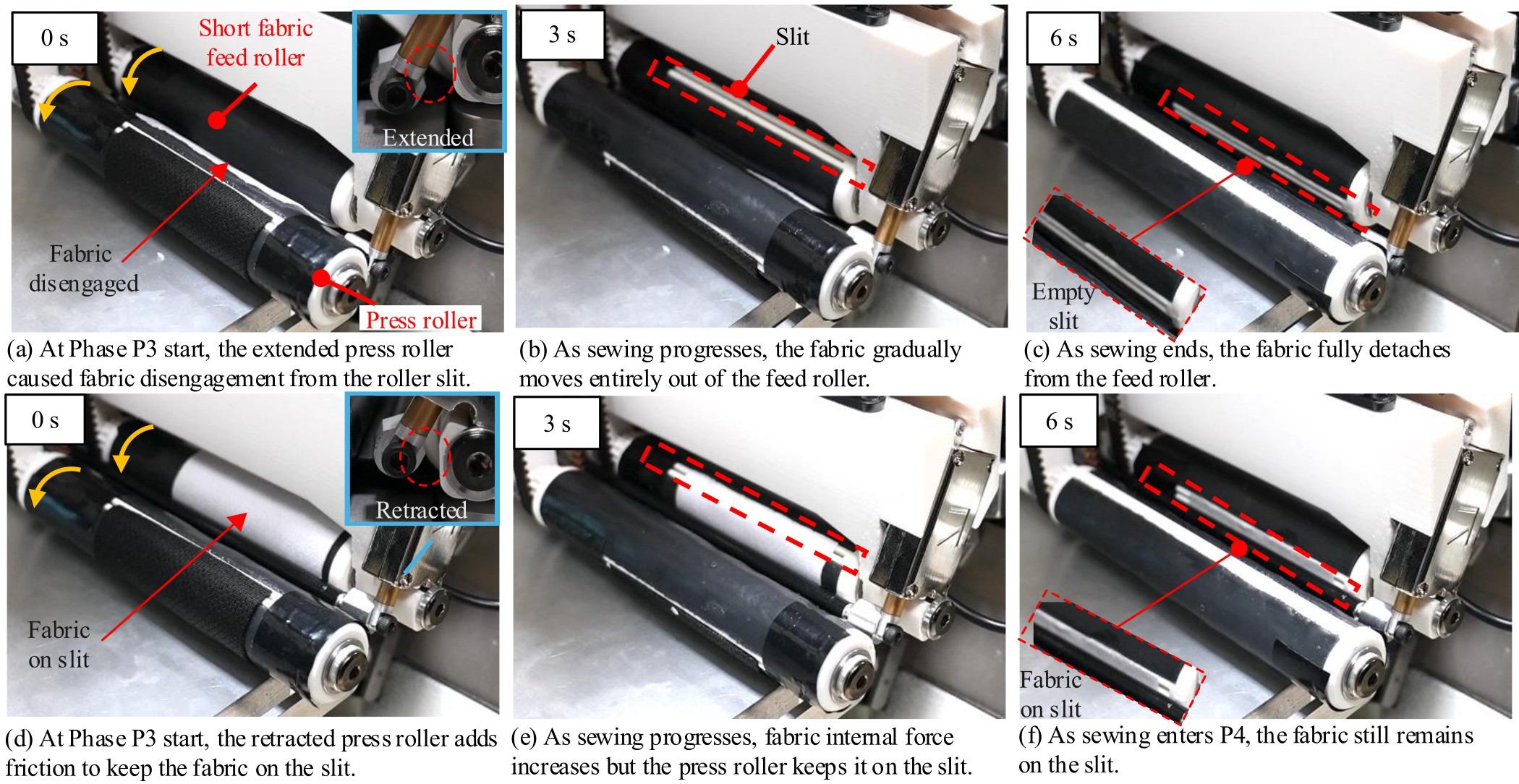}
	\caption{\ac{DLAS} Phase P3 to end experiments with press roller extended ((a)--(c)) and retracted ((d)--(e)).}
	\vspace{-1em}
	\label{fig:exp_press_roller_impact}
\end{figure*}

The monitor roller (Fig.~\ref{fig:monitor_roller}), introduced in Section~\ref{sec:monitor_roller}, measures real-time sewing speed for \ac{DLFFSC} by synchronizing its rotation with the fabric's movement. Validating its sewing speed performance with an external sensor
is impossible due to the narrow working space of the overlock
sewing machine and the deformable fabric obstructing its
view.  Hence, the sewing fabric length calculated from the roller’s speed at each \textit{sewing loop} was compared with the fixed fabric feed length of 2.6~mm per loop, with the needle reciprocating once per second.

In Fig.~\ref{fig:validating_monitor_roller}, the blue curve shows the sewn fabric length from the monitor roller angle (averaged over 6 samples), while the red dots show the length estimated from the predefined stitch length. The figure confirms that the roller’s measurement matches the sewing machine’s stitch size, indicating accurate real-time sewing speed estimation.

\subsection{DLRoSS press roller impact on {DLAS}}

\begin{figure}[t!]
	\centering
	\includegraphics[width=\figscale\columnwidth]{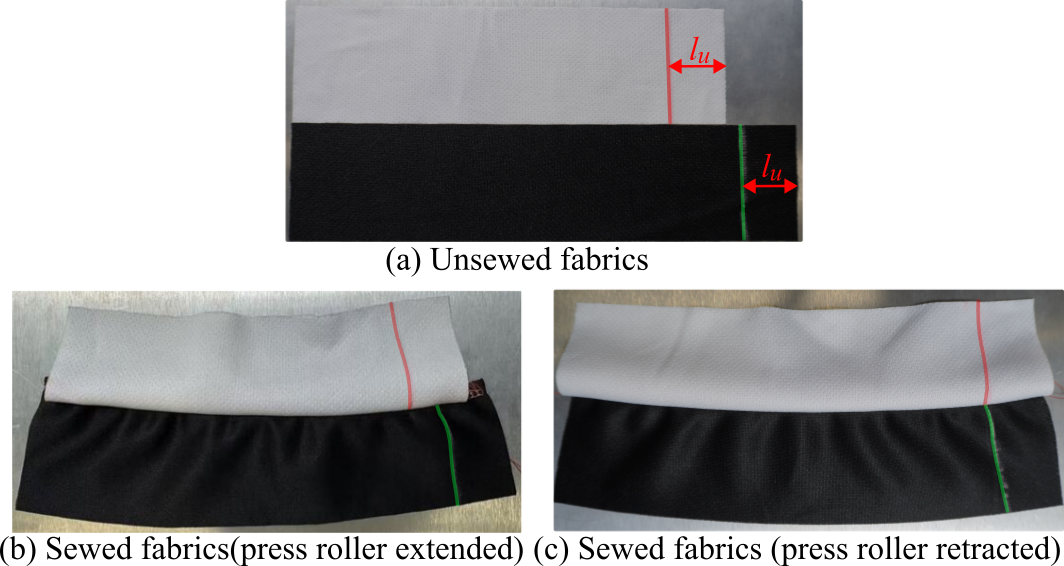}
	\caption{Fabric alignment error.}
	\vspace{-1em}
	\label{fig:uncontrollable_length}
\end{figure}

\ac{DLAS} experiments were conducted for two cases to show the importance of the \ac{DLRoSS}'s press roller: at Phase P3 the press roller was not pressed, referred to as {press roller extended} (Fig.~\ref{fig:exp_press_roller_impact} (a)--(c)), and another at P3 the press roller was pressed firmly against the fabric, referred to as {press roller retracted} (Fig.~\ref{fig:exp_press_roller_impact} (d)--(f)). 
The initial experiment setting is shown in Fig.\ref{fig:dfross}, where a white short fabric and a longer black fabric were manually aligned at one end, and placed under the sewing machine's presser foot.
On the other end, a red line was marked on one fabric and a green line on the other, each at a distance of $l_u$ from its edge (Fig.~\ref{fig:uncontrollable_length}~(a)). The distance between these two reference lines, measured along the seam line, is defined as the \textit{alignment error} and is used to evaluate performance.
Fig.~\ref{fig:uncontrollable_length} (b), and (c) show the alignment results for press roller extended, and retracted, respectively.

To evaluate the effect of press roller retraction on fabric stretching alignment error across different fabric lengths, the length of the longer fabric was kept constant at \( l_l = 200 \, \text{mm} \), while the length of the shorter fabric was varied in the experiments as \( l_s = 150 \, \text{mm} \, \text{to} \, 180 \, \text{mm} \) in increments of \( 10 \, \text{mm} \) ( \( l_s = 150:10:180 \) mm). This variation resulted in an \textit{actual fabric length ratio} of \( \gamma = l_l / l_s = 0.75:0.05:0.9 \), and in case of press roller retraction \( \rho = 0.697, 0.758, 0.818, 0.879 \), calculated from \eqref{eq:rho_updated} with \( l_{gap} = 22 \, \text{mm} \) and \( l_{slp} = 14.8 \, \text{mm} \), giving an uncontrollable fabric length of \( l_u = 36.8 \, \text{mm} \) (Fig.~\ref{fig:uncontrollable_length} (a)). For each $\gamma$, experiments were conducted with the press roller extended in all phases and then retracted at the start of Phase P3, with each set consisting of five trials.

In Fig.\ref{fig:exp_press_roller_impact} (a)--(c), during Phase P3 with the press roller extended, fabric tension caused disengagement from the roller slit. As sewing progressed, the fabric fell off the feed roller and was fully detached by the end. In Fig.\ref{fig:exp_press_roller_impact} (d)--(e), with the press roller retracted at the start of P3, added friction ensured the fabric remained on the slit, even through Phase P4.

\begin{figure}
\centering
\includegraphics[width=\figscale\columnwidth]{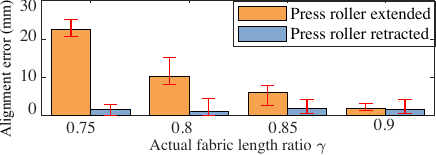}
\caption{Measured alignment error results.}
\vspace{-1em}
\label{fig:diffrho}
\end{figure}

After \ac{DLAS} ends, the sewing alignment error was measured (Fig.~\ref{fig:uncontrollable_length} (b)--(c)) for both cases, with average errors represented by orange bars (press roller extended) and blue bars (press roller retracted) in Fig.~\ref{fig:diffrho}, while the red lines indicate the maximum and minimum values.
With the press roller activated, the average sewing alignment error remains below 3~mm for all $\gamma$ values. 
For all $\gamma$, except $\gamma=0.9$, the results clearly show a significant reduction in the average sewing alignment error when the press roller was retracted (orange to blue bars, Fig.~\ref{fig:diffrho}). 
The reduction is more prominent for lower \( \gamma \), where higher fabric stretching is required, leading to greater internal fabric forces, causing the fabric slippage to occur earlier during Phase P3. However, for \( \gamma = 0.9 \), since the stretching is not eminent, the misalignment error is not significant in both cases.

\subsection{DLRoSS performance with different fabrics and lengths}

This section examines \ac{DLRoSS}'s performance in executing DLAS with varying fabric materials (Table~\ref{tab:fabric_prop}), lengths, and ratios (\( \gamma \)). The initial experimental setting is the same as that described in Sec.~V-B.

\subsubsection{\textcolor{black}{Different fabric materials}, varying $l_s$ and $\gamma$, constant $l_l$} To validate the performance of \ac{DLRoSS} in stretching shorter fabrics of varying lengths and materials, experiments were conducted with \( l_s = 150:10:180 \) mm, while keeping the longer fabric as Fabric A with \( l_l = 200 \, \text{mm} \), resulting in \( \gamma = 0.75:0.05:0.9 \). For the shorter fabric, Fabrics A, B, and C (Table~\ref{tab:fabric_prop}), each with different material properties, thickness, bending stiffness, and elastic modulus, were tested with 5 trials per fabric type per length in each set of experiments. Similar to the previous experiments, the average, maximum, and minimum sewing alignment errors were calculated as shown in Fig.~\ref{fig:new-end-effector} (a) \textcolor{black}{which shows that the average sewing alignment error for all cases is less than 3~mm. }

\begin{table}[t!]
\vspace{-1em}
\centering
\caption{Material properties of different fabrics.}
\label{tab:fabric_prop}
\renewcommand{\arraystretch}{1.2} 
\scalebox{1}{ 
    \begin{tabular}{|l|c|c|c|}
        \hline
        \textbf{Properties} & \textbf{Fabric A} & \textbf{Fabric B} & \textbf{Fabric C} \\ 
        \hline
        \multirow{-2.5}{*}{\centering Specimen}  
        & \includegraphics[width=1.6cm]{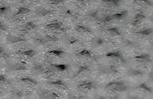}    
        & \includegraphics[width=1.6cm]{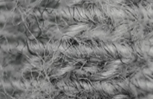}       
        & \includegraphics[width=1.6cm]{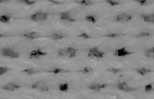}    \\
        \hline
        Material  & Polyester  & Cotton\&Nylon & Polyester \\
        \hline
        $T$ (mm) & 0.5  &  1.2   &  1.0    \\
        \hline
        $B$ (mN$\cdot$mm)   & 3.09   & 19.85    & 16.04 \\ 
        \hline
        $E$ (MPa)   & 6.23 & 4.74 & 21.34 \\ 
        \hline
        \multicolumn{4}{|l|}{\footnotesize $T$: Fabric thickness, $B$: Bending stiffness, $E$: Elastic modulus.} \\
        \hline
    \end{tabular}
} 
\vspace{-1em}
\end{table}

Although Fabric C is made of the same material as Fabric A, it is more difficult to stretch due to its high elastic modulus resulting from different weaving methods. However, irrespective of this, \ac{DLRoSS} successfully stretched the fabric for all \( \gamma \) values, except \( \gamma = 0.75 \), where further stretching was no longer possible and could cause fabric tearing. 

\subsubsection{\textcolor{black}{Different fabric materials}, varying $l_l$ and $l_s$, constant $\gamma$} 
\begin{figure}[t!]
	\centering
	\includegraphics[width=\figscale\columnwidth]{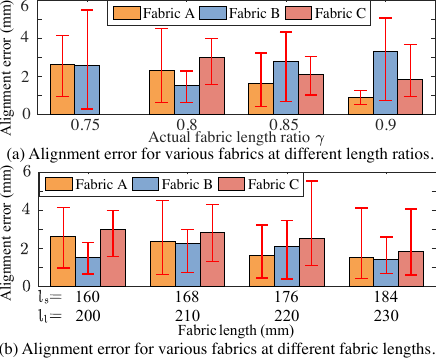}
	\caption{Measured alignment error for various fabrics.}
	\vspace{-1em}
	\label{fig:new-end-effector}
\end{figure}

To demonstrate the capability of \ac{DLRoSS} in handling different fabrics of various lengths, the fabric length ratio was kept constant at \( \gamma = 0.8 \), while varying \( l_l = 200:10:230 \) mm and \( l_s = 160:8:184 \) mm. With the longer fabric as Fabric~A, experiments were conducted using three shorter fabrics, A, B, and C, with 5 trials each. The alignment error results in Fig.~\ref{fig:new-end-effector} (b) show that the average errors remained below 3~mm.

\textcolor{black}{The sewing alignment accuracy and its standard deviation show only minor differences across fabric lengths and materials. This minimal variation is attributed to careful planning of the feed roller speed (Sec.~\ref{sec:dlffsc}) and the press roller preventing relative fabric movement, both of which ensure reliable sewing performance under diverse conditions. Across 115 trials, the maximum error remained below 6 mm, well within the JIS tolerance for shirt girth (+8 mm to \mbox{-}5 mm) \cite{JIS,9551585}.}

\section{Conclusion}
\acresetall
This letter presented the design and development of \ac{DLRoSS}, a novel robotic sewing system that automates \ac{DLAS} by stretching a shorter fabric to match a longer one and sewing them together in a straight line. 

The system features a custom end-effector with dual active feed rollers, a passive press roller for independent fabric control, and a speed-sensing unit for real-time sewing speed feedback. It executes \ac{DLAS} through four operation phases to ensure smooth and precise fabric alignment sewing.

Experiment results on different fabric sizes and materials confirmed its effectiveness. The press roller notably reduced alignment errors, achieving sub-millimeter precision. Since stretching difficulty depends on the fabric ratio rather than absolute length, experiments were performed with ratios up to 0.75, a demanding case rarely seen in industry. Future work will address curved-line sewing of stretchable fabrics for more complex garment construction for which the vision-based manipulator trajectory planning will be needed.

\addtolength{\textheight}{-1cm}   
                                  
\bibliographystyle{IEEEtran}
\bibliography{8-mybib}

\end{document}